\documentclass[fleqn,10pt]{wlscirep}
\usepackage{url}
\usepackage[utf8]{inputenc}
\usepackage[T1]{fontenc}
\usepackage{lineno}
\usepackage{multirow}
\usepackage{float}
\usepackage{enumitem}
\usepackage{soul}
\usepackage{hyperref} 
\usepackage{bm}
\hypersetup{
    colorlinks=true,
    linkcolor=blue,      
    citecolor=blue,      
    urlcolor=blue        
}
\usepackage{url}
\definecolor{darkgreen}{RGB}{4,179,1}

\title{A Leaf-Level Dataset for Soybean–Cotton Detection and Segmentation}

\author[1,*]{Thiago H. Segreto}
\author[1]{Juliano Negri}
\author[1]{Paulo H. Polegato}
\author[1]{João Manoel Herrera Pinheiro}
\author[1]{Ricardo V. Godoy}
\author[1]{Marcelo Becker}
\affil[1]{Mechanical Engineering Department, S\~ao Carlos School of Engineering, University of S\~ao Paulo, S\~ao Carlos 13566-590, SP, Brazil}

\affil[*]{corresponding author(s): Thiago H. Segreto (thiago.segreto.silva@alumni.usp.br)}

\begin{abstract}

Soybean and cotton are major drivers of many countries’ agricultural sectors, offering substantial economic returns but also facing persistent challenges from volunteer plants and weeds that hamper sustainable management. Effectively controlling volunteer plants and weeds demands advanced recognition strategies that can identify these amidst complex crop canopies. While deep learning methods have demonstrated promising results for leaf-level detection and segmentation, existing datasets often fail to capture the complexity of real-world agricultural fields. To address this, we collected 640 high-resolution images from a commercial farm spanning multiple growth stages, weed pressures, and lighting variations. Each image is annotated at the leaf-instance level, with 7,221 soybean and 5,190 cotton leaves labeled via bounding boxes and segmentation masks, capturing overlapping foliage, small leaf size, and morphological similarities. We validate this dataset using YOLO11, demonstrating state-of-the-art performance in accurately identifying and segmenting overlapping foliage. Our publicly available dataset supports advanced applications such as selective herbicide spraying and pest monitoring and can foster more robust, data-driven strategies for soybean–cotton management.

\end{abstract}

\begin{document}

\flushbottom
\maketitle


\section*{Background \& Summary}
    \label{sec:introduction}

Soybean (\textit{Glycine max (L.) Merr.}) and cotton (\textit{Gossypium hirsutum L.}) are among the most significant crops in many countries’ agriculture, playing a key role in Brazil's national economy~\cite{conab2024, fao2023}. 
The integration of these crops into rotational systems, such as soybean–cotton or soybean–corn, enhances soil health and fertility while mitigating agronomic risks commonly associated with monoculture practices~\cite{culturaalgodao2017}. However, this rotation introduces specific management challenges, particularly the emergence of volunteer plants, soybean or cotton shoots that germinate from residual seeds of previous seasons. These compete directly with the main crop for critical resources, including water, light, and nutrients, and can serve as hosts for pests such as the cotton boll weevil (\textit{Anthonomus grandis}), posing a significant threat to crop yield and necessitating rigorous management strategies~\cite{agrolink2023,imamt2019}.

Traditional management of weeds or unwanted volunteer plants predominantly relies on blanket herbicide spraying. Although this approach is often effective, its overuse has led to significant concerns, including the emergence of herbicide-resistant weed populations, rising input costs, and adverse environmental impacts~\cite{cavenaghi2012,ikeda2010}. In parallel, the management of residual cotton stalks remains critical for effective pest control, especially in disrupting the life cycle of the cotton boll weevil. This practice is not only agronomically essential but also legally mandated in regions where the sanitary void policy enforces the eradication of volunteer plants to prevent pest proliferation~\cite{arruda2022,menin2017}. These challenges corroborate the critical need for innovative and precise control strategies capable of accurately distinguishing between volunteer plants, active crops, and weeds, thereby enabling more targeted and sustainable interventions~\cite{imamt2019,manual2012}.

Recent computer vision and deep learning advances have introduced robust solutions for automating plant identification, contributing to more sustainable agricultural practices. Convolutional Neural Networks (CNNs) and Transformer-based vision architectures, particularly when applied to semantic segmentation and object detection, can classify foliage at multiple scales, distinguishing crops and weeds even in complex field conditions~\cite{Wu2021,Silva2024,Sunil2024}. 
While semantic segmentation excels in pixel-level detail, facilitating tasks such as green cover estimation, leaf area indexing, and early disease detection~\cite{Wu2021}, bounding-box detection offers a more computationally efficient alternative for large-scale mapping and plant identification.
When employed together, these methods provide both broad spatial context and refined morphological analyses~\cite{champApplicationInPlantSciences}. Indeed, prior studies have successfully employed segmentation and detection to differentiate weeds from crops such as carrot~\cite{czymmek2019vision}, cotton~\cite{chen2022performance}, and sugarcane~\cite{yano2016identification}. However, many existing datasets still focus on a single task (often weed detection) and lack comprehensive annotation formats, e.g., semantic masks without instance labels or only bounding boxes, limiting the ability of deep learning models to manage overlapping crops and diverse weed pressures~\cite{madec2023vegann}.

Despite the importance of robust training data, comprehensive real-world agricultural datasets, particularly those supporting instance segmentation, remain scarce. For example, the Moving Fields Weed Dataset~\cite{Genze2024} provides numerous annotations but is constrained to controlled indoor environments. In contrast, Champ et al.~\cite{champApplicationInPlantSciences} captured outdoor scenes yet included only about 700 labels per target crop (2,489 total annotations), while the Leaf Segmentation and Counting Challenge~\cite{MINERVINI201680} offers just 284 labeled instances. Many existing resources also focus on narrow applications, such as disease detection or single-species classification, thus limiting the adaptability of models to diverse field conditions~\cite{Zenkl2022}. Although datasets like~\textit{DeepWeeds} address multiple weed species~\cite{Wu2021}, they often lack the multi-modal annotations required for complex tasks, including precise herbicide application and morphological trait analysis~\cite{Fan2022}.

\begin{figure}[t!]
\centering
\includegraphics[width=\linewidth]{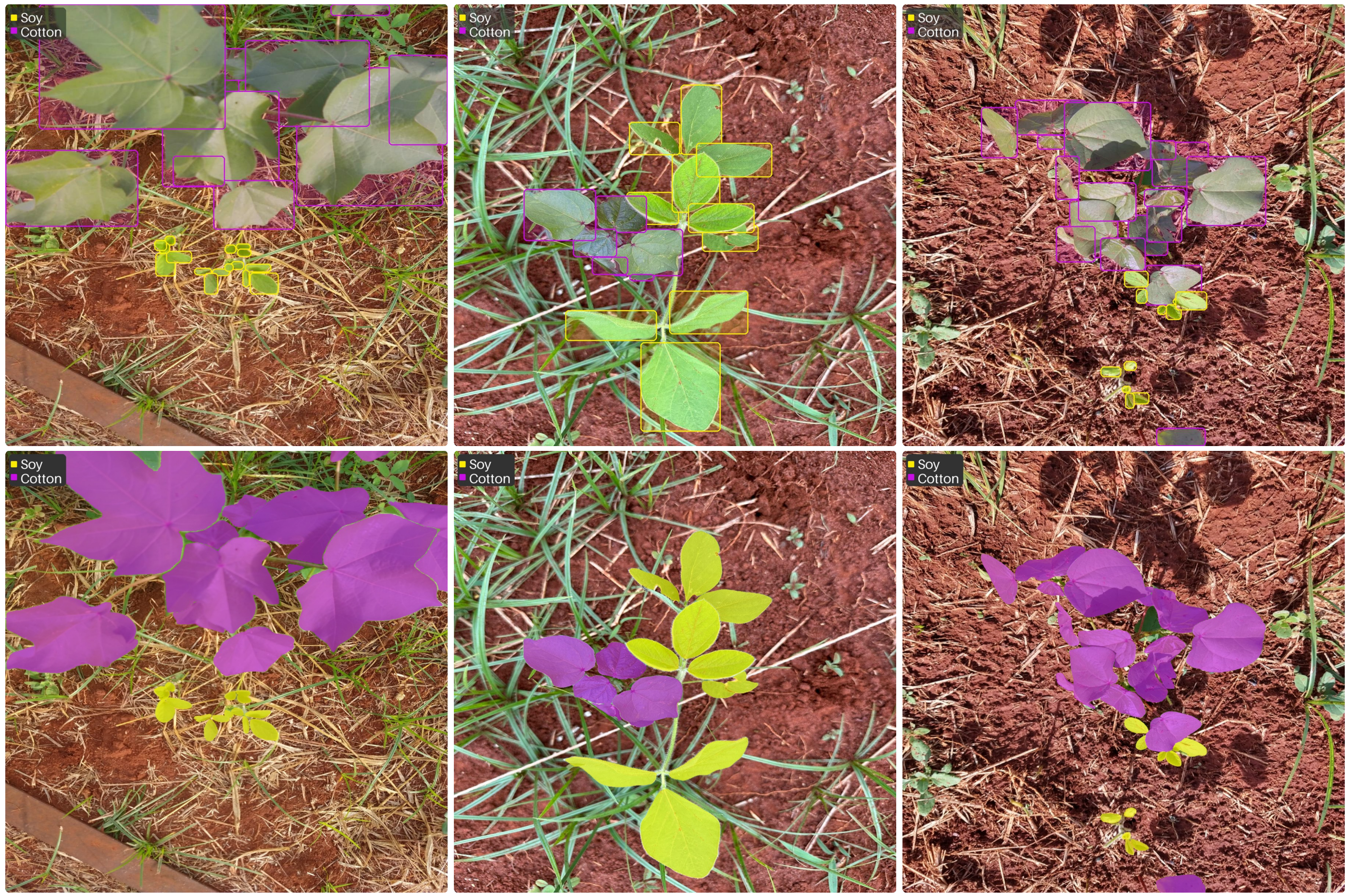}
\caption{Ground-truth annotations include detection bounding boxes, shown in the first row, and segmentation masks, shown in the second row.
}
\label{fig:segXod}
\end{figure}

To meet this need, in this paper, we introduce a soybean–cotton leaf detection and segmentation dataset developed under realistic field conditions on a farm in São Paulo, Brazil. The dataset features images across different growth stages, environmental lighting conditions, and degrees of weed pressure, which are known requirements to make deep learning applications robust in the outdoor uncontrolled scenario~\cite{woebbecke1995color, gongal2015sensors}. Each image is meticulously annotated with both soybean and cotton leaves, even when partial occlusions or overlapping foliage present significant identification challenges. By accommodating both object detection and semantic segmentation formats, as shown in Fig.~\ref{fig:segXod}, this resource enables advanced applications: (1) mapping the locations soybean or cotton leaves for targeted herbicide spraying, (2) distinguishing the active crop from surrounding weeds through relatively simple computational filters, and (3) extracting precise morphological features for phenotyping or disease surveillance~\cite{Zenkl2022,Wu2021}.

\section*{Methods}
    \label{sec:methods}

\begin{figure}
    \centering
    \includegraphics[width=0.75\linewidth]{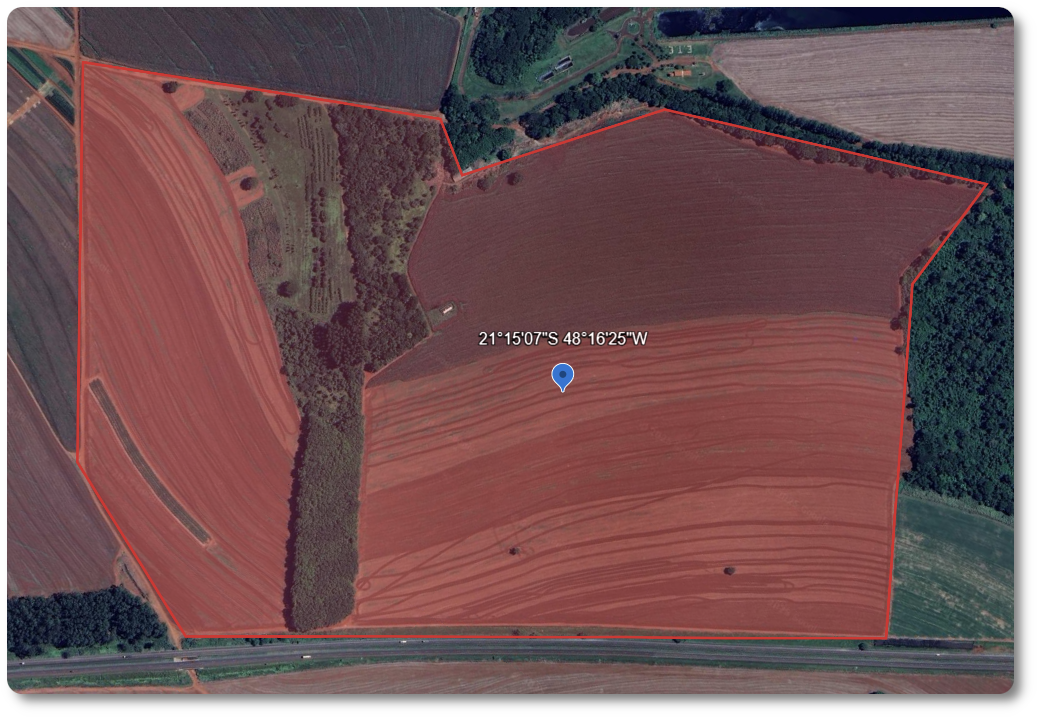}
    \caption{Satellite imagery of the data collection site in Jaboticabal, São Paulo, Brazil. The red polygon delineates the farm's specific boundaries where image acquisition took place. Maps data: Google and Airbus\@ 2025}
    \label{fig:farm}
\end{figure}

\paragraph{Image Acquisition.}
All images were collected on a farm located in Jaboticabal (S\~ao Paulo, Brazil) over a ten-week period. \autoref{fig:farm} shows the exact location of the collection. 
Field personnel, composed of three experts, captured approximately 70 images weekly, from morning to late afternoon (from 6\,am 
to 6\,pm) to capture a diverse range of light conditions. This yielded a total of 700 images. However, an automated quality-control procedure discarded 60 low-quality or near-duplicate samples, leaving 640 high-quality photos.

\begin{figure}[t!]
\centering
\includegraphics[width=\linewidth]{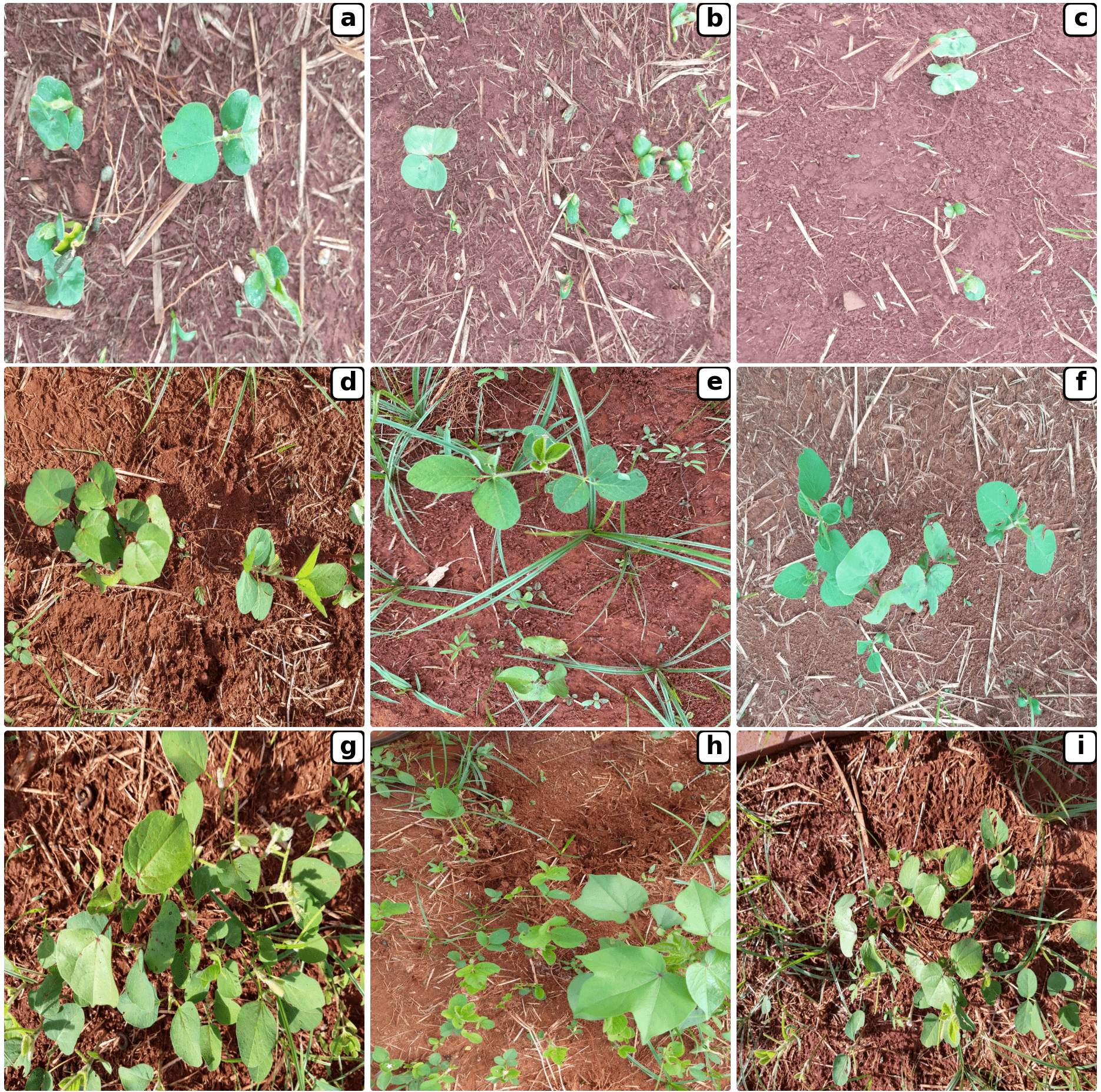}
\caption{Growth stage variations in soybean and cotton fields. A 3$\times$3 grid of raw images illustrates 
early (a--c), middle (d--f), and dense (g--i) canopy stages. In the early stage (1--3 weeks), 
sparse foliage and minimal leaf overlap simplify segmentation but offer limited complexity. The middle stage (4--7 weeks) 
introduces denser coverage, partial occlusions, and moderate weed presence. The dense stage (8--10 weeks) exhibits 
substantial leaf overlap, shading, and varied leaf sizes, posing increased challenges for both detection and segmentation.}
\label{fig:grid_3x3_images}
\end{figure}

Images were taken at roughly knee-to waist-level using personal smartphone cameras, all set to 1600\,$\times$\,1200 resolution. Despite the consistent output resolution, the differences in camera models and vantage points were intentional, reflecting the diversity of real-world acquisition scenarios. Soybean fields ranged from emergence (VE) to near full seed (R6), while cotton fields progressed from emergence to a pre-boll stage, collectively providing a broad spectrum of phenotypic appearances. Notably, no herbicides were applied during the collection period, allowing various weed species to remain in the background. This choice ensured a realistic context where the target crops coexisted with non-target vegetation. To further illustrate the progression of crop development across the ten-week span, Fig.~\ref{fig:grid_3x3_images} presents a 3$\times$3 grid of representative raw images capturing early, middle, and dense foliage stages in both soybean and cotton fields. 

\paragraph{Annotation Process.}
Leaf-level annotations were generated using the Computer Vision Annotation Tool (CVAT)~\cite{CVAT2023}, 
aided by the Segment Anything Model (SAM)~\cite{kirillov2023segany}. Two experts specializing in soybean 
and cotton identification performed the core labeling, with a third reviewer overseeing consistency checks. 
Any leaf recognizable as soybean or cotton, despite potential occlusions or slight blur, was annotated with 
an instance segmentation mask and a corresponding bounding box to support both semantic segmentation 
and object detection tasks. 

\paragraph{Annotation Criteria.}
\begin{itemize}[noitemsep, topsep=2.0pt, left=2.0pt]
    \item \textbf{Inclusion:} Every visible soybean or cotton leaf was labeled, regardless of size, health, or overlap with other foliage, provided experts reached a consensus on its class.
    \item \textbf{Exclusion:} Weeds, soil, and other background elements were \emph{not} annotated (though they remain unaltered in the images). Leaves too obscured for definite class determination were also excluded.
\end{itemize}

\begin{figure}[t!]
\centering
\includegraphics[width=\linewidth]{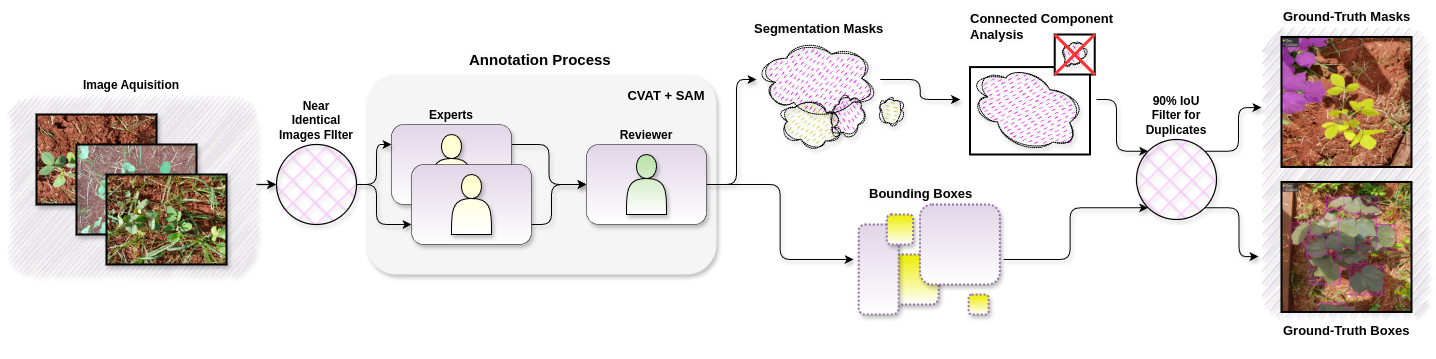}
\caption{Illustration of the dataset creation and annotation workflow. Field images are first acquired under near-vertical perspectives and filtered to remove near-identical samples. Experts and a reviewer then generate initial segmentation masks and bounding boxes in CVAT, assisted by the SAM. Connected component analysis eliminates small “blob” artifacts in the masks, and any duplicate labels are merged using a 90\% IoU filter. The final output includes precise ground-truth masks and bounding boxes for soybean and cotton leaves.
}
\label{fig:annotation_pipeline}
\end{figure}

This procedure yielded 12{,}411 annotated leaves. 7{,}221 soybean and 5{,}190 cotton, and given the large volume of annotated soybean and cotton leaves, occasional human labeling errors led to duplicate or near-duplicate entries. To mitigate this, we filtered duplicate annotations by applying a 90\% Intersection over Union (IoU) threshold, effectively removing redundant labels that could undermine data integrity. In parallel, we discovered random ''pixel blob'' artifacts introduced by the Segment Anything Model (SAM). Although often subtle, these blobs could obscure true leaf boundaries and were more noticeable on blurrier masks. Leveraging OpenCV~\cite{opencv_library}, we performed connected component analysis to detect and remove small, disconnected 
blobs, thereby refining the precision of each segmentation mask. Fig.~\ref{fig:annotation_pipeline} showcases the entire dataset creation pipeline.

\begin{table}[ht]
    \centering
    \caption{Summary of Dataset Composition}
    \label{tab:dataset_stats}
    \begin{tabular}{lcc}
        \toprule
        \textbf{Parameter} & \textbf{Value} & \textbf{Notes} \\
        \midrule
        \textbf{Total Images}       & 640              & After discarding 60 low-quality samples \\
        \textbf{Resolution}         & 1600\,$\times$\,1200 & Multiple smartphone cameras \\
        \textbf{Soybean Leaves}     & 7{,}221          & From emergence (VE) to near R6 \\
        \textbf{Cotton Leaves}      & 5{,}190          & From emergence to pre-boll stage \\
        \textbf{Annotations}        & Masks + bounding boxes & CVAT with SAM \\
        \bottomrule
    \end{tabular}
\end{table}

Table~\ref{tab:dataset_stats} provides a concise overview of the final dataset, including the number of annotated leaves for each crop, imaging resolution, and annotation formats. Because no herbicides were applied, weeds and other background elements remained in the fields, offering 
a realistic context for assessing algorithmic robustness. The deliberate variability in viewing angles, lighting conditions, and growth stages aims to challenge both segmentation and detection algorithms, while the presence of diverse weed types adds an additional layer of complexity. In this regard, this dataset enables tasks ranging from basic crop\st{/weed} discrimination to advanced leaf-level trait extraction.

\paragraph{Automated Annotation Paradigms.} While our methodology relied on expert verification aided by segmentation models, the broader field of data annotation is increasingly integrating Large Language Models (LLMs) and Vision-Language Models (VLMs) as "labeling copilots" to scale dataset creation~\cite{tan2024large}. In these workflows, VLMs facilitate zero-shot auto-labeling and multimodal grounding for visual tasks~\cite{carion2025sam3segmentconcepts}, effectively generating bounding boxes and class descriptions, while LLMs can generate synthetic training data~\cite{sapkota2025zeroshotautomaticannotationinstance}. This paradigm shifts the primary human role from manual instance creation to high-level instruction and programmatic quality control, significantly reducing annotation latency. However, despite the efficiency of these foundation models, they introduce challenges regarding hallucinations and inherent biases, necessitating robust human-in-the-loop adjudication to ensure the trustworthiness required for high-precision agricultural applications. Therefore, the use of LLMs and VLMs will be assessed in future works.

\section*{Data Records}
    \label{sec:data_records}


The dataset \cite{Segreto} in this study are publicly available under a \textit{CC\_BY\_4.0} license at:
\begin{center}
    {\textbf{\href{https://doi.org/10.6084/m9.figshare.28466636.v3}}{https://doi.org/10.6084/m9.figshare.28466636.v3}.}
\end{center}
The repository contains a single top-level folder named \texttt{SoyCotton}, which is organized into two subfolders:

\begin{itemize}
    \item \texttt{images}: This folder contains all of the RGB images used in the dataset. Each file is saved in PNG format at a resolution of 1600$\times$1200 pixels.
    \item \texttt{annotations}: This folder holds the \texttt{coco.json} file, which follows the standard COCO format and provides both bounding-box and instance-segmentation labels for soybean and cotton leaves. 
\end{itemize}

Within the \texttt{annotations} file, each image is referenced by a unique identifier, and every annotated leaf is associated with its corresponding bounding box coordinates and polygonal mask. Users can leverage common COCO-compatible libraries or toolkits to parse, visualize, and modify these annotations as needed.

\section*{Technical Validation}
    \label{sec:technical}

\label{sec:techincal}
This section outlines the procedures and results used to validate the dataset for both object detection and instance segmentation tasks. We first detail the rationale for selecting our validation model, followed by a description of the data splitting strategy, cross-validation methodology, and the specific metrics used to evaluate model performance against the dataset's annotations.

\begin{figure}[t!]
\centering
\includegraphics[width=\linewidth]{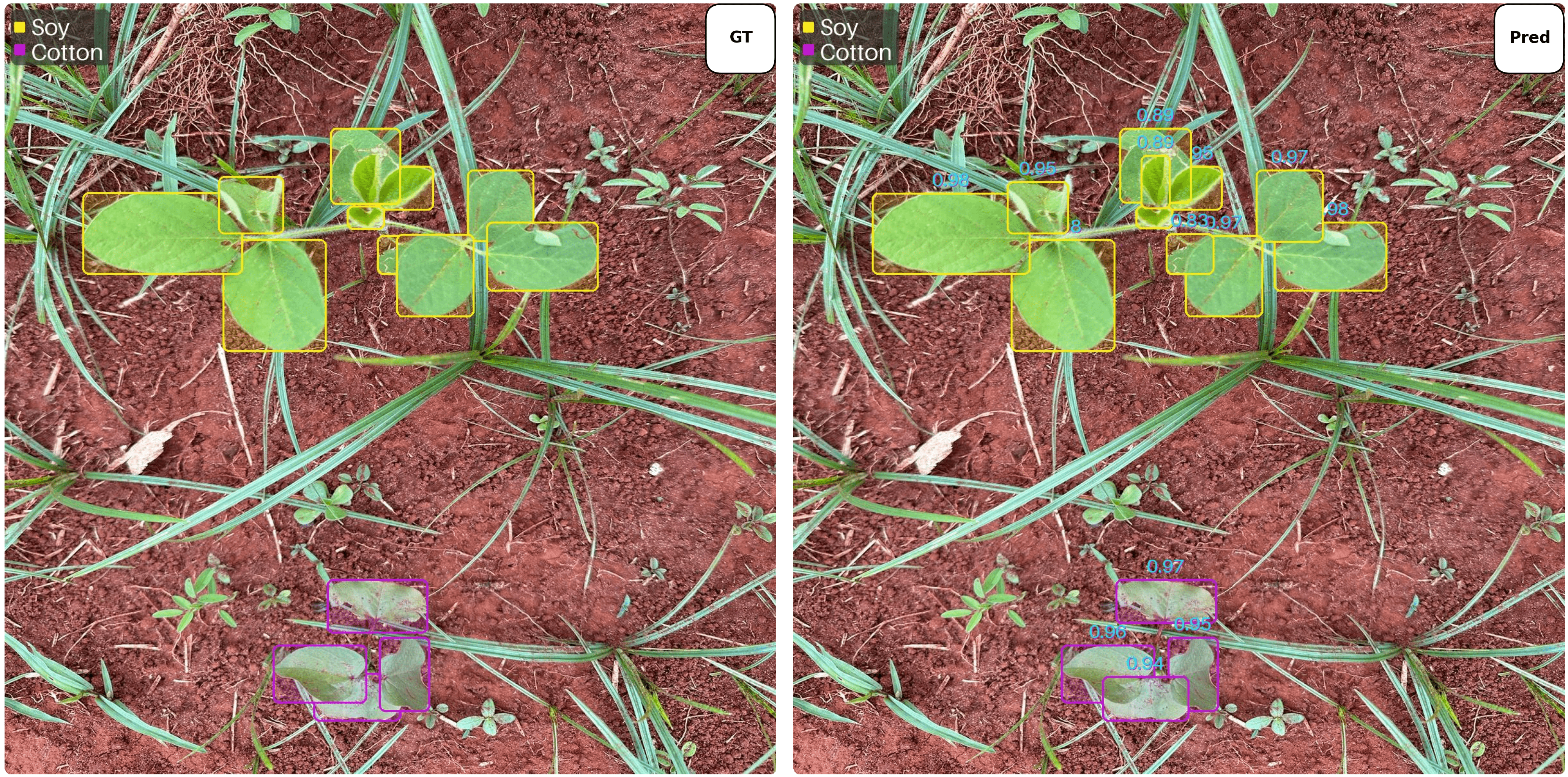}
\caption{Detection comparison. The left image (GT) depicts ground-truth bounding boxes for soybean (yellow) and cotton (purple). 
The right image (Pred) shows the model’s bounding-box outputs with confidence scores (blue).}
\label{fig:detection_comparison}
\end{figure}

\paragraph{Dataset Splits.}
All data splits described below were shuffled and stratified by soybean and cotton class ratios using scikit-learn~\cite{scikit-learn}, 
ensuring representative distributions across training, validation, and test sets.

\begin{itemize}
    \item \textbf{Hyperparameter Tuning (80--10--10):} The dataset was divided into 80\% for training, 10\% for validation, and 10\% for a hold-out test set. This split provided a balanced foundation for tuning model parameters and performing initial performance checks.

    \item \textbf{Five-Fold Performance Analysis (80--20):} From the same 80\% training portion, we applied a five-fold cross-validation strategy to further evaluate model stability. Each fold used 80\% of the training set for training and 20\% for validation, mitigating overfitting to a single split.

    \item \textbf{Data Ablation Study (90--10):} A separate experiment allocated a fixed 10\% of the dataset for testing. The remaining 90\% was subdivided into ten equal chunks (each 9\% of the full data) using \texttt{StratifiedShuffleSplit} from \texttt{scikit-learn}. This procedure ensured that the distribution of soybean and cotton instances, as well as the proportion of images representing different canopy densities (a proxy for growth stages), remained consistent across every data increment. This stratification mirrors the original dataset's composition, ensuring that performance gains are attributable to dataset size rather than variations in sample difficulty.
\end{itemize}

\paragraph{Classification Metrics: Precision, Recall, and F1-Score.}  
For evaluating the classification performance of object detection and segmentation, we use precision, recall, and the F1-score, defined as follows:  
\[
\text{Precision} = \frac{\text{TP}}{\text{TP} + \text{FP}}, \quad \text{Recall} = \frac{\text{TP}}{\text{TP} + \text{FN}}, \quad \text{F1} = 2 \times \frac{\text{Precision} \times \text{Recall}}{\text{Precision} + \text{Recall}}.
\]  
\textit{Precision} quantifies the accuracy of positive predictions, representing the proportion of predicted instances (e.g., detected objects or segmented regions) that are correct. A high precision indicates fewer false positives, meaning the model avoids over-predicting.  
\textit{Recall}, also known as sensitivity, measures the model’s ability to detect all actual instances, expressing the fraction of ground-truth objects or regions successfully identified. High recall implies fewer missed detections (i.e., fewer false negatives).  
The \textit{F1-score} is the harmonic mean of precision and recall, providing a single metric that balances the trade-off between the two. It is particularly useful when the model’s performance must account for both false positives and false negatives equally, favoring neither over-detection nor under-detection.

\begin{figure}[t!]
\centering
\includegraphics[width=\linewidth]{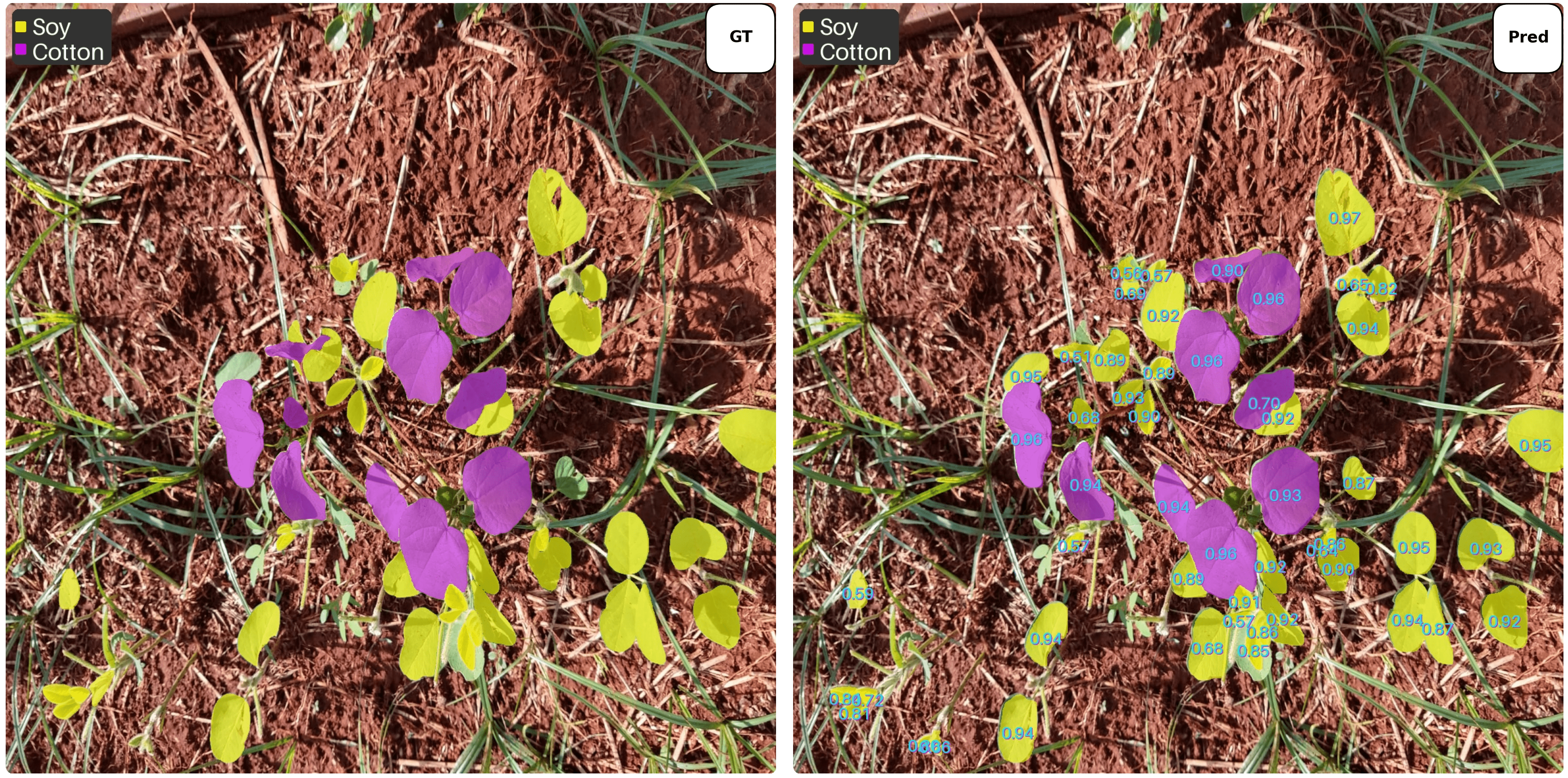}
\caption{Segmentation comparison. The left image (GT) illustrates manual annotations, with soybean leaves 
in yellow and cotton leaves in purple, whereas the right image (Pred) presents the model-generated segmentation 
masks.}
\label{fig:seg_comp}
\end{figure}

\paragraph{Intersection over Union (IoU).}  
IoU is a fundamental metric used to assess the spatial overlap between a predicted bounding box or mask and its corresponding ground-truth annotation. It is defined as:  
\[
\text{IoU} = \frac{\text{Area of Intersection}}{\text{Area of Union}},
\]  
where the ''Area of Intersection'' is the overlapping region between the predicted and ground-truth areas, and the ''Area of Union'' is the total area spanned by both, excluding double-counting of the intersection. IoU ranges from 0 to 1, with 1 indicating a perfect match and 0 indicating no overlap. In object detection and segmentation, a prediction is typically classified as a true positive if its IoU with the ground truth exceeds a specified threshold (e.g., 0.5). IoU serves as the foundation for computing mean Average Precision (mAP), as it determines the correctness of each prediction.

\paragraph{Mean Average Precision at IoU=0.5 (\(\text{mAP}_{50}\)).}  
The \(\text{mAP}_{50}\) metric assesses performance at an IoU threshold of 0.5. For each class \(c\), predictions are ranked by confidence scores, and precision and recall are computed at each rank. The Average Precision (AP) for class \(c\) is the area under the Precision–Recall curve, often approximated using the all-point interpolation method:  
\[
\text{AP}_{50}^c = \sum_{k=1}^{N_c} \left( \text{Recall}_{k}^c - \text{Recall}_{k-1}^c \right) \cdot \max_{j \geq k} \text{Precision}_{j}^c,
\]  
where \(N_c\) is the number of predictions for class \(c\), \(\text{Recall}_{k}^c\) and \(\text{Precision}_{k}^c\) are the recall and precision at rank \(k\), and \(\text{Recall}_{0}^c = 0\). The \(\text{mAP}_{50}\) is then the mean of AP across all \(C\) classes:  
\[
\text{mAP}_{50} = \frac{1}{C} \sum_{c=1}^{C} \text{AP}_{50}^c.
\]  
An IoU threshold of 0.5 implies that a prediction is correct if it overlaps at least 50\% with the ground truth, making \(\text{mAP}_{50}\) a relatively lenient measure of localization accuracy.

\paragraph{Mean Average Precision at IoU=0.5:0.95 (\(\text{mAP}_{50-95}\)).}  
The \(\text{mAP}_{50-95}\) metric extends \(\text{mAP}_{50}\) by averaging \(\text{mAP}\) scores over IoU thresholds from 0.5 to 0.95 in steps of 0.05 (i.e., \(T = \{0.5, 0.55, 0.6, \ldots, 0.95\}\), totaling 10 thresholds). For each threshold \(t \in T\) and class \(c\), \(\text{AP}_{t}^c\) is computed as above, adjusting the TP/FP/FN classification based on \(t\). The \(\text{mAP}_{50-95}\) is then:  
\[
\text{mAP}_{50-95} = \frac{1}{|T|} \sum_{t \in T} \left( \frac{1}{C} \sum_{c=1}^{C} \text{AP}_{t}^c \right) = \frac{1}{10} \sum_{t=0.5}^{0.95, \Delta t=0.05} \text{mAP}_{t},
\]  
where \(\text{mAP}_{t} = \frac{1}{C} \sum_{c=1}^{C} \text{AP}_{t}^c\) is the mAP at threshold \(t\), and \(|T| = 10\). This metric rigorously evaluates localization precision, as higher IoU thresholds (e.g., 0.95) demand near-perfect overlap, making \(\text{mAP}_{50-95}\) a robust indicator of model quality across varying levels of spatial accuracy.

\paragraph{Detection and Segmentation Model.} 
We selected the YOLO11 model for this study based on its specific advantages for real-time agricultural instance segmentation. While two-stage models, such as Mask R-CNN~\cite{he2017mask}, offer high precision, their significant computational overhead and sequential architecture make them challenging for real-time deployment in the field~\cite{kotthapalli2025survey}. Among the one-stage models, the YOLO family provides the necessary real-time inference capabilities~\cite{jeghamh2025benchmark}. We specifically chose YOLO11 over other recent variants (e.g., YOLOv8~\cite{yolov8}, YOLOv10~\cite{wang_yolov10_2024}) for two key reasons. First, YOLO11 offers a unified, state-of-the-art framework that natively supports both the bounding-box detection and instance segmentation tasks required by our dataset~\cite{khanam2024yolov11, kotthapalli2025survey, sapkota2025ultralytics}. Second, recent comprehensive benchmarks demonstrate that the YOLO11 family achieves a superior balance of accuracy and computational efficiency compared to many of its predecessors~\cite{jeghamh2025benchmark}, and achieves accuracy on par or superior to other state-of-the-art models~\cite{YOLO11_ultralytics,wang_yolov10_2024} such as RT-DETR.

\begin{table}[t!]
    \centering
    \caption{Comparative performance analysis of mainstream detection and segmentation models on the SoyCotton dataset. Inference times were measured on an NVIDIA L40S GPU (FP16, Batch=1). Best results in each category are highlighted in bold.}
    \label{tab:model_comparison}
    \begin{tabular}{lcccc}
        \toprule
        \textbf{Model} & \textbf{mAP\textsubscript{50}} & \textbf{mAP\textsubscript{50-95}} & \textbf{F1-Score} & \textbf{Inference (ms)} \\
        & {\scriptsize ↑} & {\scriptsize ↑} & {\scriptsize ↑} & {\scriptsize ↓} \\
        \midrule
        \multicolumn{5}{c}{\textit{Detection Models}} \\
        YOLOv8m & 0.837 & 0.728 & 0.801 & \textbf{4.76} \\
        YOLOv9m & 0.855 & 0.728 & 0.811 & 7.53 \\
        YOLOv10m & 0.844 & 0.719 & 0.799 & 5.69 \\
        RT-DETR-L & 0.847 & 0.726 & \textbf{0.820} & 14.97 \\
        \textbf{YOLO11m} & \textbf{0.862} & \textbf{0.741} & 0.818 & 6.27 \\
        \midrule
        \multicolumn{5}{c}{\textit{Segmentation Models}} \\
        YOLOv8m-seg & 0.890 & 0.702 & 0.849 & \textbf{4.87} \\
        YOLOv9c-seg & \textbf{0.896} & 0.704 & \textbf{0.853} & 7.41 \\
        FastSAM-x & 0.893 & \textbf{0.713} & 0.851 & 7.33 \\
        \textbf{YOLO11m-seg} & 0.895 & 0.712 & \textbf{0.853} & 6.10 \\
        \bottomrule
    \end{tabular}
\end{table}

We conducted comparative experiments against other well-established architectures (including YOLOv8, YOLOv9, YOLOv10, and RT-DETR) with their default hyperparameters to assess how they performed for our dataset. All inference times were measured on a single NVIDIA L40S GPU (batch size 1, FP16 precision) to eliminate data-loading overhead. For object detection, we compared YOLO11m against its predecessor YOLOv8m, the newer YOLOv9m and YOLOv10m, and the Transformer-based RT-DETR-L. For instance segmentation, we evaluated YOLO11m-seg against YOLOv8m-seg, YOLOv9c-seg, and the segment-anything-based FastSAM-x. The results are shwon in Table~\ref{tab:model_comparison}.

In the detection task, YOLO11m achieved the highest accuracy ($0.862$ mAP\textsubscript{50} and $0.741$ mAP\textsubscript{50-95}). In terms of inference speed, while YOLOv8m was the fastest ($4.76$ ms), YOLO11m remained highly competitive at $6.27$ ms. notably, the Transformer-based RT-DETR-L exhibited higher latency ($14.97$ ms), nearly 2.5 times that of YOLO11m.

In the segmentation task, YOLO11m-seg achieved the highest F1-score ($0.853$). While YOLOv9c-seg achieved a marginally higher mAP\textsubscript{50} ($0.896$ vs $0.895$), YOLO11m-seg offered a superior balance of speed and accuracy. Although YOLOv8m-seg ($4.87$ ms) was the fastest, YOLO11m-seg ($6.10$ ms) outperformed both FastSAM-x ($7.33$ ms) and YOLOv9c-seg ($7.41$ ms) in inference speed. This confirms that YOLO11m-seg retains robust real-time capabilities while delivering on-par classification performance.

\paragraph{Hyperparameter Optimization.}
Standard YOLO11 configurations are calibrated for large-scale, general-purpose datasets (e.g., COCO) and may not be the most optimal when applied to small, custom datasets with distinct domain shifts. Agricultural environments present unique challenges, such as high foliage density, heavy occlusion, and homogeneous textures, that can differ from standard object detection benchmarks. Furthermore, depending on leaf age and the camera viewpoint, leaves can appear very small, further complicating detection and segmentation tasks~\cite{liu_deep_2019}. To adapt the model to these specific constraints, we optimized the medium-sized YOLO11 architecture using a variant of the \textit{Tree-structured Parzen Estimator (TPE)}~\cite{bergstra_making_nodate}, a Bayesian approach known for effective hyperparameter tuning in diverse deep learning applications~\cite{watanabe_tree-structured_2023}.


We aimed to produce hyperparameter settings suitable for both detection and segmentation. Thus, we applied the \textit{Multiobjective TPE (MOTPE)}~\cite{ozaki_multiobjective_2022} via Optuna~\cite{akiba_optuna_2019} to optimize F1-score (capturing classification performance) and mAP\textsubscript{50-95} (measuring spatial accuracy) simultaneously. To expedite the search process, we used distributed training with ClearML~\cite{clearml} across multiple workers, coupled with the Adam optimizer~\cite{kingma2014adam}.

To quantify the impact of this optimization, we benchmarked the final TPE-tuned model against the default YOLO11 configuration. As shown in Table~\ref{tab:optimization_impact}, the optimized hyperparameters provided \textbf{7\%} improvement in mAP\textsubscript{50-95} for both detection and segmentation. This improvement in high-IoU metrics confirms that the TPE-based search was very useful to boost performance on this custom dataset.

\begin{table}[t!]
    \centering
    \caption{Impact of MOTPE hyperparameter optimization on model performance. The comparison demonstrates the gains achieved by the optimized model over the default YOLO11 configuration (Baseline).}
    \label{tab:optimization_impact}
    \begin{tabular}{llccc}
        \toprule
        \textbf{Task} & \textbf{Configuration} & \textbf{mAP\textsubscript{50}} & \textbf{mAP\textsubscript{50-95}} & \textbf{F1-Score} \\
        \midrule
        \multirow{2}{*}{Detection} & Default & 0.862 & 0.740 & 0.817 \\
         & \textbf{Optimized} & \textbf{0.905} & \textbf{0.812} & \textbf{0.858} \\
        \midrule
        \multirow{2}{*}{Segmentation} & Default & \textbf{0.896} & 0.712 & \textbf{0.856} \\
         & \textbf{Optimized} & 0.892 & \textbf{0.786} & 0.833 \\
        \bottomrule
    \end{tabular}%
\end{table}

Table~\ref{tab:tpe_hparams} presents the 11 hyperparameters most critical to convergence, regularization, and data augmentation. Each parameter differs from its default setting to maximize performance gains under the challenging conditions posed by highly similar crop foliage.

\begin{table}[t!]
\centering
\caption[Optimal YOLOv11 medium model hyperparameters]{Optimal hyperparameters derived via MOTPE for the YOLOv11 medium-sized model, optimized for F1 and mAP\(_{50-95}\). Each column indicates the final best value found for \textit{Detection} vs.\ \textit{Segmentation}, and the final column contains the \textit{Default} values of the model, differing from default settings.}
\label{tab:tpe_hparams}
\begin{tabular}{lccc}
\toprule
\textbf{Hyperparameter} & \textbf{Detection} & \textbf{Segmentation} & \textbf{Default} \\
\midrule
Initial Learning Rate (\texttt{LR0})       & 0.0001 & 0.0046 & 0.01 \\
Learning Rate Factor (\texttt{LRF})       & 0.095  & 0.097  & 0.01 \\
Momentum (\texttt{MOMENTUM})             & 0.92   & 0.80   & 0.93 \\
Weight Decay (\texttt{WEIGHT\_DECAY})    & 0.0001 & 0.0003 & 0.0005 \\
Box Loss Gain (\texttt{BOX})             & 6.0    & 6.5    & 7.5 \\
Classification Loss Gain (\texttt{CLS})  & 1.0    & 0.1    & 0.5 \\
DFL Loss Gain (\texttt{DFL})             & 0.3    & 2.3    & 1.5 \\
Mixup Augmentation (\texttt{MIXUP})      & 0.3    & 0.1    & 0 \\
Scale Augmentation (\texttt{SCALE})      & 0.9    & 0.6    & 0.5 \\
Perspective Augmentation (\texttt{PERSPECTIVE}) & 0.001 & 0.0 & 0 (max 0.001)  \\
Translate Augmentation (\texttt{TRANSLATE}) & 0.1    & 0.0 & 0.1 \\
\bottomrule
\end{tabular}
\end{table}

\paragraph{Five-Fold Cross-Validation Performance on YOLO11 Medium.}
Following the hyperparameter search, the YOLO11 medium model was subjected to a five-fold cross-validation for both detection (bounding boxes) and segmentation (instance masks). 
Tables~\ref{tab:kfold_detection} and \ref{tab:kfold_segmentation} summarize the overall precision (P), recall (R), F1-score, 
\(\text{mAP}_{50}\), and \(\text{mAP}_{50-95}\) across folds for detection and segmentation, respectively. Each row includes the mean 
and standard deviation over the five folds, offering insight into performance variability.

\begin{table}[H]
\centering
\caption{Five-fold cross-validation results for YOLO11 medium on \textbf{detection}. 
P, R, and F1 are reported as mean $\pm$ standard deviation, 
along with $\text{mAP}_{50}$ and $\text{mAP}_{50-95}$.}
\label{tab:kfold_detection}
\begin{tabular}{l c c c c c}
\toprule
\textbf{Category} & \textbf{P} & \textbf{R} & \textbf{F1} & \(\textbf{mAP}_{50}\) & \(\textbf{mAP}_{50-95}\) \\
\midrule
All    & 0.883 $\pm$ 0.005 & 0.835 $\pm$ 0.005 & 0.858 $\pm$ 0.001 & 0.905 $\pm$ 0.001 & 0.812 $\pm$ 0.003 \\
Soy    & 0.858 $\pm$ 0.009 & 0.803 $\pm$ 0.009 & 0.830 $\pm$ 0.001 & 0.877 $\pm$ 0.001 & 0.770 $\pm$ 0.003 \\
Cotton & 0.907 $\pm$ 0.004 & 0.867 $\pm$ 0.007 & 0.887 $\pm$ 0.003 & 0.934 $\pm$ 0.002 & 0.854 $\pm$ 0.004 \\
\bottomrule
\end{tabular}
\end{table}

\begin{table}[H]
\centering
\caption{Five-fold cross-validation results for YOLO11 medium on \textbf{segmentation}.}
\label{tab:kfold_segmentation}
\begin{tabular}{l c c c c c}
\toprule
\textbf{Category} & \textbf{P} & \textbf{R} & \textbf{F1} & \(\textbf{mAP}_{50}\) & \(\textbf{mAP}_{50-95}\) \\
\midrule
All    & 0.864 $\pm$ 0.010 & 0.804 $\pm$ 0.009 & 0.833 $\pm$ 0.006 & 0.892 $\pm$ 0.004 & 0.786 $\pm$ 0.005 \\
Soy    & 0.829 $\pm$ 0.020 & 0.774 $\pm$ 0.009 & 0.800 $\pm$ 0.008 & 0.863 $\pm$ 0.006 & 0.748 $\pm$ 0.004 \\
Cotton & 0.899 $\pm$ 0.014 & 0.834 $\pm$ 0.015 & 0.865 $\pm$ 0.009 & 0.923 $\pm$ 0.004 & 0.824 $\pm$ 0.007 \\
\bottomrule
\end{tabular}
\end{table}

\begin{figure}[ht]
\centering
\includegraphics[width=\linewidth]{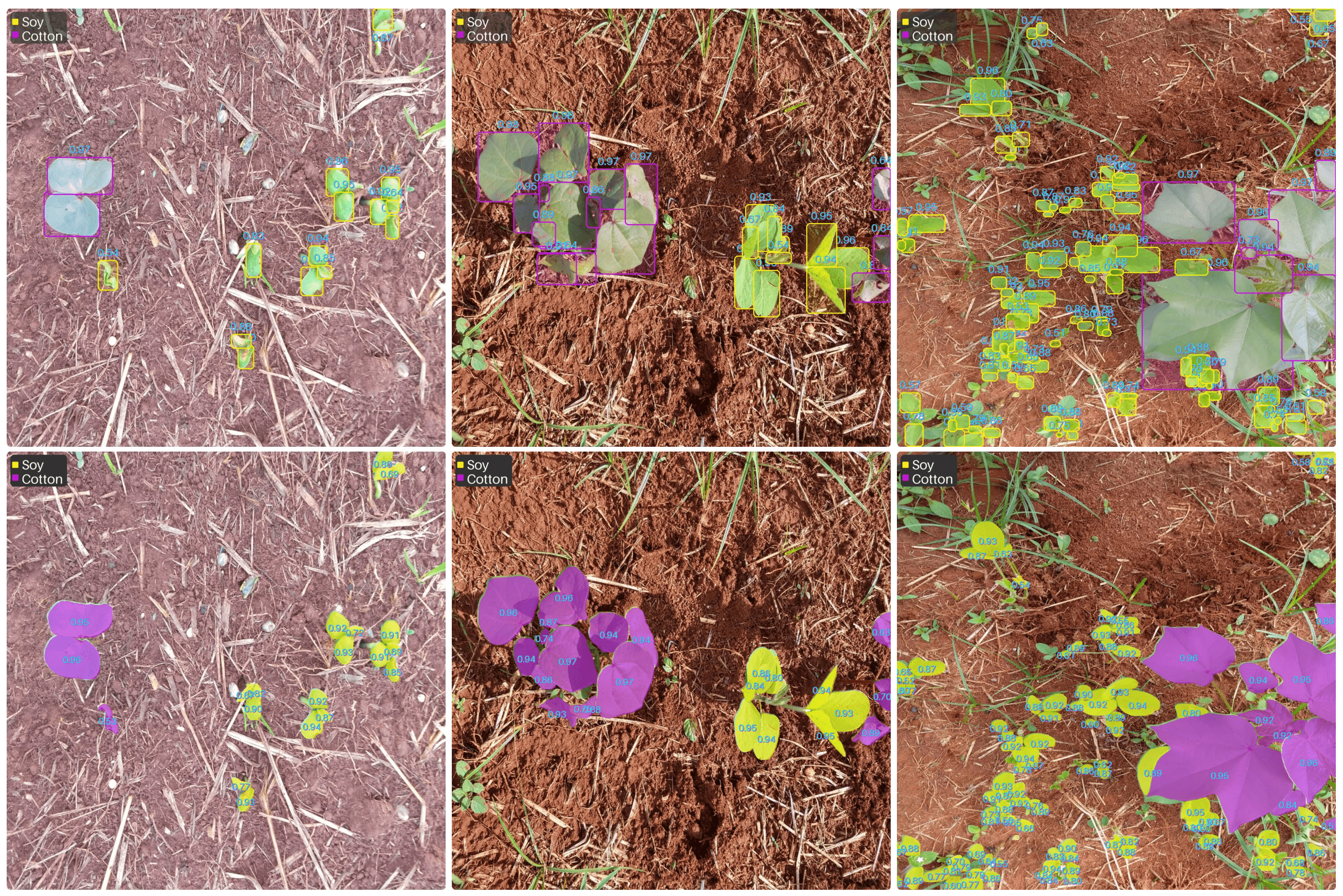}
\caption{Detection and segmentation across soybean and cotton growth stages. This 2\(\times\)3 grid displays YOLO11m outputs: the top row shows predicted bounding boxes, and the bottom row presents corresponding segmentation masks for early, mid, and dense leaf maturity (left to right). Early-stage predictions exhibit clear leaf separation, while dense-stage outputs reflect challenges posed by occlusions and overlapping foliage.}
\label{fig:predictions}
\end{figure}

Figures~\ref{fig:detection_comparison} and~\ref{fig:seg_comp} provide complementary visualizations. 
Figure~\ref{fig:detection_comparison} highlights the model's detection predictions alongside ground-truth bounding boxes, 
showing its capacity to localize soybean and cotton leaves accurately. Meanwhile, Figure~\ref{fig:seg_comp} compares ground-truth 
segmentation masks with the model's predicted masks, revealing how well YOLO11 medium delineates leaf boundaries 
at the pixel level. The numbers associated with the bounding boxes and segmentation masks in the prediction image represent the classification confidence values.

Detection yields slightly higher precision, recall, and F1-score than segmentation, due to 
the simpler task of predicting bounding boxes compared to the complexity of pixel-wise annotation. 
As seen in Tables~\ref{tab:kfold_detection} and~\ref{tab:kfold_segmentation}, the ``All'' category achieves a mean 
precision of 0.883 for detection versus 0.864 for segmentation. This gap is also mirrored in recall (0.835 vs.\ 0.804) 
and F1-score (0.858 vs.\ 0.833). Such differences are expected, as delineating precise boundaries in densely populated 
images often demands higher variability in training samples and more intricate feature extraction.

\begin{figure}[t!]
\centering
\includegraphics[width=\linewidth]{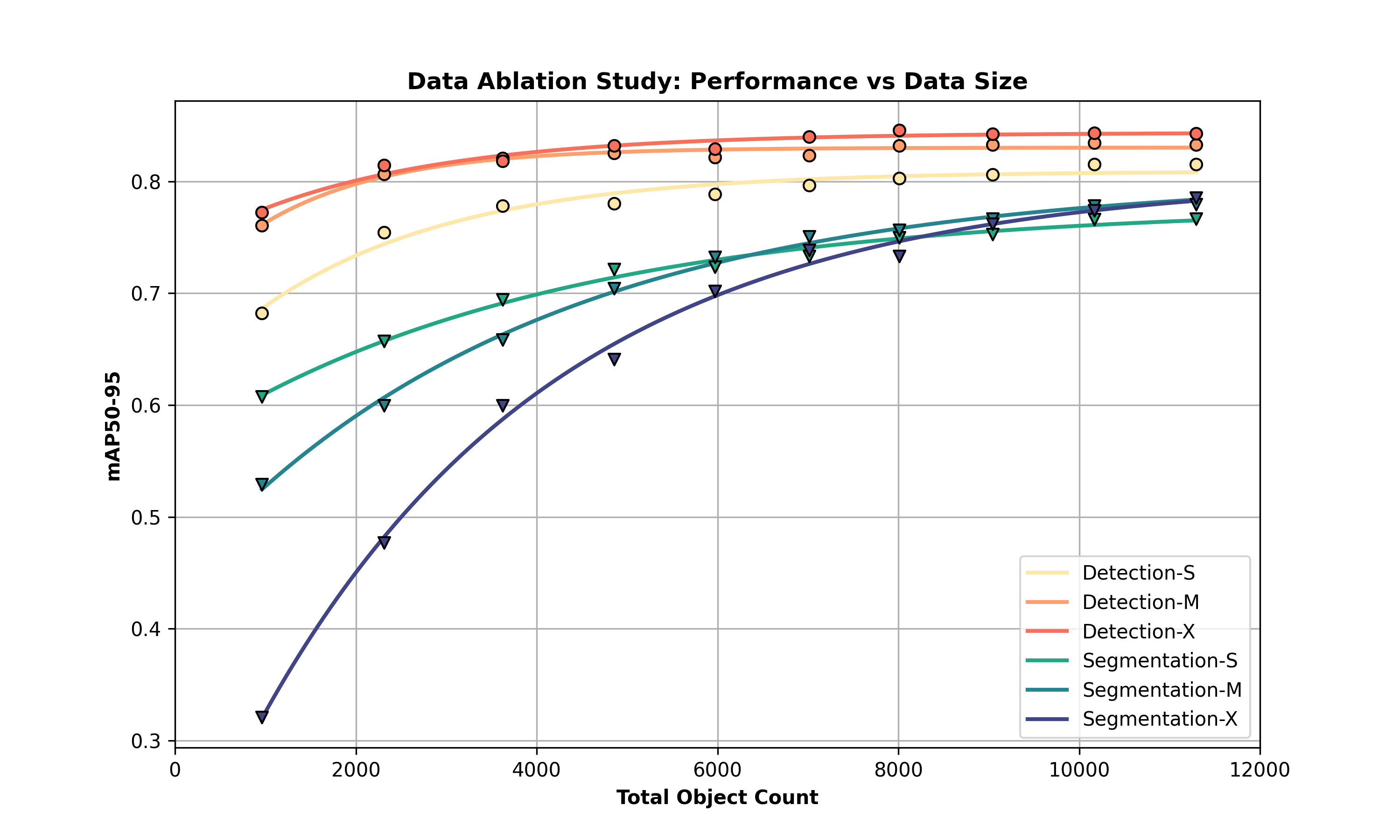}
\caption[Data size ablation study]{Effect of dataset size on YOLOv11 performance for detection (yellow for S, orange for M, red for X) and segmentation (green for S, dark green for M, dark blue for X). The x-axis indicates the total number of annotated objects used in training, and the y-axis shows \(\text{mAP}_{50-95}\). Detection appears to plateau near 7,000 objects, while segmentation gains persist until roughly 9,000--11,000 objects. An exponential curve is fitted to the data to illustrate the saturation effect.}
\label{fig:data_ablation}
\end{figure}

\paragraph{Performance analysis.}
In terms of crop-specific performance, cotton exhibits marginally higher metrics than soybean across both tasks. 
For detection, cotton's precision of 0.907 and recall of 0.867 surpass soybean's 0.858 and 0.803, respectively. 
The trend persists in segmentation metrics as well. In order to counteract this performance gap, we included extra images with more soybean leaves in the dataset. However, even with more than 2,000 annotated soybean leaves compared to cotton, soybean still outperformed cotton. 

Lastly, Fig.~\ref{fig:predictions} showcases predictions at different growth stages: early, mid, and dense canopies. 
In the early phase, with fewer overlapping leaves, bounding boxes and segmentation masks tend to align closely with 
ground truth. However, as foliage density increases (e.g., near canopy closure), performance on both tasks degrades 
due to heightened occlusion and partial leaf coverage. For robust field-level applications, these findings emphasize 
the importance of training on images representing diverse plant densities and stages.


\paragraph{Data Ablation Analysis.}
To investigate the impact of dataset size on model performance, we conducted a data ablation analysis using three variants of the YOLO11 model, small (S), medium (M), and large (X), trained on incrementally increasing portions of the dataset. The dataset~\cite{Segreto}, comprising annotated cotton and soybean objects, was divided using a 90--10 split: 90\% of the data was allocated for training and partitioned into equal increments, while the remaining 10\% was reserved as a fixed test set to ensure consistent evaluation across all experiments. The training increments were stratified and randomly sampled to preserve class balance at each ablation step.

Figure~\ref{fig:data_ablation} illustrates the relationship between \(\text{mAP}_{50-95}\) and the number of annotated objects for each model--task combination. An exponential curve was fitted to these data points via \texttt{curve\_fit} in \textit{scipy} to capture the saturation trend better as dataset size increases. We note that detection tasks tend to plateau around 7,000 objects, whereas segmentation tasks often continue to improve until reaching approximately 9,000--11,000 objects. This discrepancy indicates that bounding-box detection saturates earlier, while pixel-level segmentation benefits from additional annotations, likely due to its higher granularity and complexity.

Table~\ref{tab:marginal_gains} provides a mock-up for quantifying marginal improvements in \(\text{mAP}_{50-95}\) at key annotation thresholds. Although all model variants (S, M, and X) show strong early gains, the rate of improvement drops markedly once the plateau region is reached. Notably, the YOLO11-X model, with 56.9 million parameters, retains an edge over smaller variants, especially in segmentation tasks, suggesting that a higher capacity model can better leverage a larger dataset.


\begin{table}[t!]
\centering
\caption[Marginal model gains with dataset size]{Marginal improvements in detection and segmentation performance at three data-size milestones for the YOLOv11 medium model.}
\label{tab:marginal_gains}
\begin{tabular}{lccccc}
\hline
\textbf{Annotated Objects} & \textbf{Det. mAP$_{50-95}$} & \textbf{Seg. mAP$_{50-95}$} & \textbf{Det. Gain} & \textbf{Seg. Gain} \\
\hline
9,042  & 83.3\% & 76.6\% & \textbf{\textcolor{darkgreen}{+0.1\%}}    & \textbf{\textcolor{darkgreen}{+1.0\%}} \\
10,167 & 83.5\% & 77.8\% & \textbf{\textcolor{darkgreen}{+0.2\%}}    & \textbf{\textcolor{darkgreen}{+1.2\%}} \\
11,293 & 83.3\% & 78.0\% & \textbf{\textcolor{red}{-0.2\%}}    & \textbf{\textcolor{darkgreen}{+0.1\%}} \\
\hline
\end{tabular}
\end{table}


\begin{table}[t!]
\centering
\caption{Computational cost vs. final performance trade-off for YOLO11 variants (S, M, X) on the full training dataset. mAP results are extracted from the final data point of the ablation study. Compute metrics are from the official model benchmarks.}
\label{tab:cost_performance}
\begin{tabular}{llccc}
\hline
\textbf{Task} & \textbf{Model} & \textbf{Parameters (M)} & \textbf{GFLOPS (B)} & \textbf{Final mAP$_{50-95}$ (\%)} \\
\hline
\textbf{Detection} & YOLO11-S & 9.4 & 21.5 & 81.6 \\
 & YOLO11-M & 20.1 & 68.0 & 83.3 \\
 & YOLO11-X & 56.9 & 194.9 & 84.4 \\
\hline
\textbf{Segmentation} & YOLO11-S & 10.1 & 33.0 & 77.0 \\
 & YOLO11-M & 22.4 & 113.2 & 78.0 \\
 & YOLO11-X & 62.1 & 296.4 & 78.8 \\
\hline
\end{tabular}
\end{table}

This analysis highlights two practical considerations:
 
 \begin{itemize}
    \item \textbf{Early vs. Late Plateau:} If the primary goal is rapid bounding-box detection, annotating beyond 7,000 objects yields only marginal gains. Conversely, segmentation requires fine-grained annotations and benefits from additional data up to around 9,000-11,000 objects.
    \item \textbf{Model Complexity vs. Performance Trade-off:} A secondary consideration is the trade-off between model size and accuracy. Table ~\ref{tab:cost_performance} compares the computational cost (parameters and GFLOPS) of the YOLO11 small (S), medium (M), and large (X) variants against their final mAP$_{50-95}$ performance on our dataset (using the full 11.3k training objects). For detection, the YOLO11-X model, which is 2.8x larger than the -M model in parameters (56.9M vs. 20.1M) and GFLOPS, offered only a marginal 1.1\% mAP gain (84.4\% vs. 83.3\%). A similar trend occurred in segmentation, where the -X model provided almost no discernible mAP improvement over the -M model (78.8\% vs. 78.0\%) despite a 2.6x increase in computational cost. This demonstrates a point of diminishing returns, reinforcing that the YOLO11-M model (used in our 5-fold validation) provides a strong balance between computational efficiency and accuracy for this specific agricultural dataset.
 \end{itemize}

Future work might investigate how advanced data augmentation or alternative labeling methods could further lift performance, especially once the model begins to plateau. As such, beyond these observed thresholds, increasing annotation volume alone may not guarantee performance gains. Refinements in data quality, model architecture, or training strategy could be more impactful.

\paragraph{Error Analysis.}
While metrics such as \(\text{mAP}\) provide a high-level view of model performance, they often hide the specific failure modes that are critical for agricultural robotics, such as the distinction between overlooking a plant (missed detection) and hallucinating one (false positive). To address this, we conducted a discretized error analysis across three growth stages, with equal time lengths within the ten-week period, for both object detection and instance segmentation, and then we categorized errors into three distinct types:

\begin{itemize}
    \item \textbf{False Negative:} A ground truth leaf that the model failed to predict with sufficient overlap (\(\text{IoU} < 0.5\)).
    \item \textbf{False Positive:} A prediction that did not overlap with any ground truth leaf (e.g., background debris or soil classified as a crop).
    \item \textbf{Misclassification:} A leaf correctly localized but assigned the wrong class label (e.g., a cotton leaf predicted as soybean).
\end{itemize}

\begin{figure}[t!]
    \centering
    \includegraphics[width=\linewidth]{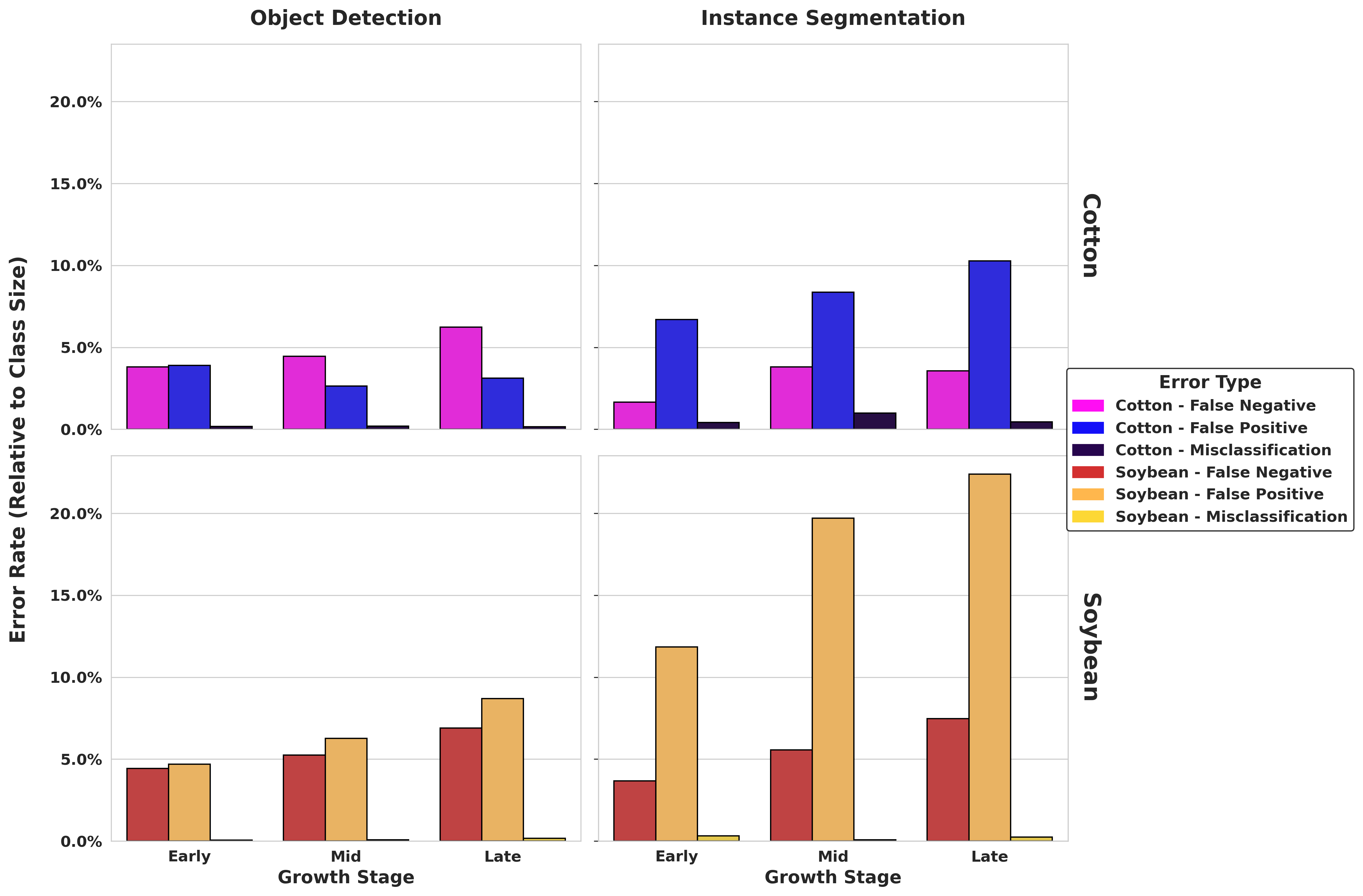}
    \caption{\textbf{Discretized error analysis across growth stages.} The bar charts compare normalized error rates for Object Detection (left) and Instance Segmentation (right). While misclassification (purple) remains negligible, Instance Segmentation suffers from higher false positive rates (light blue/orange), particularly for Soybean in the Late stage ($22.4\%$), likely due to its morphological similarity to background weeds.}
    \label{fig:error_analysis}
\end{figure}

To ensure comparability between classes with different occurrences, we normalized all error counts by the total number of ground truth instances for that class. Figure~\ref{fig:error_analysis} illustrates these error rates across the Early, Mid, and Late growth stages.

In the Early stage, the model demonstrated high classification accuracy with minor misclassification rates ($< 1\%$ for both Cotton and Soybean across both tasks). For object detection, errors were balanced between missed detections (approx.  4\%) and false positives (approx. 4\%). This suggests that at the seedling stage, the main challenge is distinguishing small leaves from the background rather than differentiating between crop types.

Moving to the Mid and Late stages, we observed a distinct divergence in error trends. Missed detections steadily increased with crop maturity. For example, cotton missed detections in object detection rose from 3.8\% (Early) to 6.2\% (Late). This trend is consistent with increased canopy closure. Figure~\ref{fig:grid_3x3_images} shows how the canopy gets denser as the plant grows, hence worse occlusion makes it difficult for the model to resolve individual leaves deep in the foliage. Conversely, misclassification rates remained remarkably low ($< 1\%$) even in the Late stage, indicating that the model has learned to effectively differentiate between soy and cotton.

\begin{figure}[t!]
    \centering
    \includegraphics[width=\linewidth]{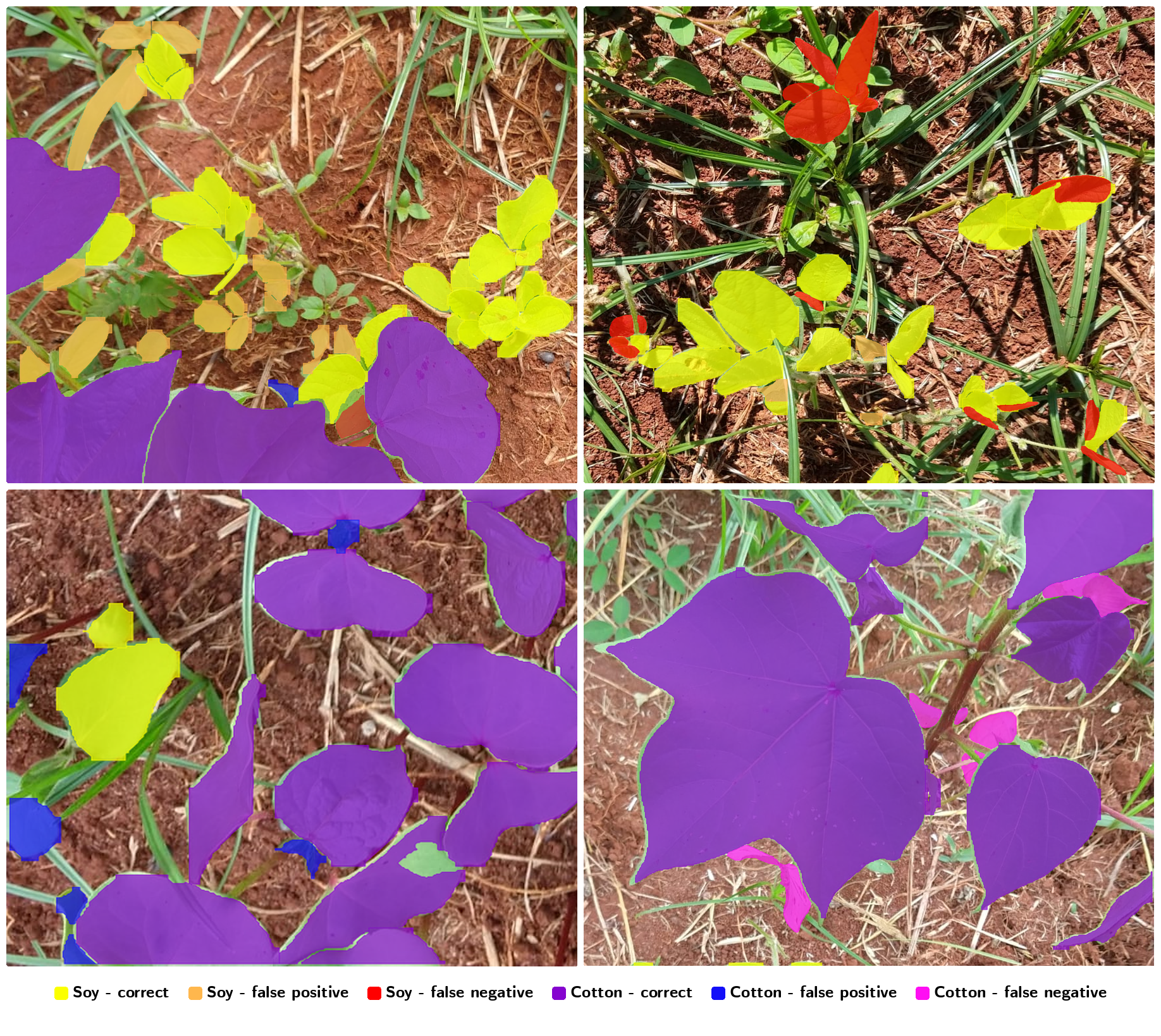}
    \caption{\textbf{Qualitative analysis of critical failure modes.} The top row illustrates the most significant false positive and false negative cases for soybean, while the bottom row displays the corresponding errors for cotton. Consistent with the quantitative findings, errors are primarily driven by severe leaf-on-leaf occlusion in dense canopies and morphological similarities between soybean foliage and background weeds.}
    \label{fig:error_grid}
\end{figure}

A performance difference was observed between tasks. Instance Segmentation consistently exhibited much higher false positive rates than Object Detection. In the Late stage, the false positive rate for Soybean reached 22.4\% for segmentation, compared to only 8.7\% for object detection. As shown in Fig.~\ref{fig:error_grid}, this is largely attributable to the inherent difficulty of segmentation. Conversely, the false negatives are primarily driven by physical challenges rather than mask precision. As illustrated in Fig.~\ref{fig:error_grid}, high leaf density leads to severe occlusion, which, when combined with localized motion blur, causes the model to overlook overlapping instances.

\section*{Code availability}
    \label{sec:code}

All scripts used to reproduce the dataset splits, as well as convert them to the YOLO format, are available at \href{https://github.com/Segretto/SoyCotton-Data}{SoyCotton-Repository}. 
This repository contains instructions in the ReadMe for replicating our experimental results and testing configurations. 
For users who prefer custom data manipulation, we recommend referring to the COCO annotation format guidelines 
to integrate these annotations into alternative pipelines.



\section*{Acknowledgements}

This work was supported by the \textit{Fundação de Apoio à Física e à Química} (FAFQ) and funded by \textit{Instituto Matogrossense do Algodão} (IMAmt) and the \textit{Cooperativa Mista de Desenvolvimento do Agronegócio} (COMDEAGRO) under grants EMBRAPII PIFS-2111.0043.


\section*{Author contributions statement}

    T.H.S.: Conceptualization, Methodology, Data Curation, Validation, Writing - J.N.: Data Curation, Validation, Writing \& Editing - P.H.P.: Methodology, Data Curation, Writing \& Editing - J.M.H.P.: Methodology, Validation - R.G.: Methodology, Writing \& Editing - M.B.: Project Administration, Resources, Supervision - 
    All authors reviewed the manuscript. 

\section*{Competing interests}

    The authors declare no competing interests.

\newpage
\bibliography{main.bib}
\end{document}